\documentclass{article}

 \usepackage[dblblindworkshop, final]{neurips_2026}
\workshoptitle{Foundation Models for Temporal Systems}

\usepackage[utf8]{inputenc} 
\usepackage[T1]{fontenc}    
\usepackage{hyperref}       
\usepackage{url}            
\usepackage{booktabs}       
\usepackage{amsfonts}       
\usepackage{nicefrac}       
\usepackage{microtype}      
\usepackage{xcolor}         
\usepackage{graphicx}
\usepackage{multirow}
\usepackage{rotating}
\usepackage{amsmath}
\usepackage{amssymb}
\usepackage{amsthm}
\usepackage{bm}
\usepackage{marvosym}

\title{Interweaving Marginals into Multivariate Sample Paths: Training-Free Dependence Construction for Probabilistic Time Series Foundation Models}

\author{
  Jinmyeong Choi\qquad Jinkwan Jang\qquad Seul Lee\qquad Taesup Kim\textsuperscript{\Letter} \\ 
  Graduate School of Data Science \\ 
  Seoul National University \\ 
  \texttt{\{jinmyeongchoi, jkjang22, ds\_seul, taesup.kim\}@snu.ac.kr}
}

\begin{document}

\maketitle

\begin{abstract}
Probabilistic time series foundation models (TSFMs) provide coordinate-wise predictive distributions, but these marginals do not determine a joint distribution over multivariate future trajectories. We evaluate deliberately simple, training-free dependence baselines built from established Gaussian copulas and historical rank coupling. Our contribution is a controlled empirical study: every channel--horizon coordinate keeps exactly the same marginal sample multiset across methods, isolating the effect of coupling. Across 14 tasks and three frozen backbones, these simple historical baselines improve targeted temporal and cross-channel dependence diagnostics. Energy Score effects are smaller and vary by method and budget. The targeted improvements persist at 10, 50, and 100 paths under both controlled coupling and direct sampling, including native multivariate backbone inference. These findings establish an empirical reference for evaluating richer dependence models over frozen TSFM marginals.
\end{abstract}

\section{Introduction}
\label{sec:introduction}

Many probabilistic time series foundation models (TSFMs) can provide marginal
predictive quantiles for multiple channels and horizons without task-specific
retraining \citep{ansari2025chronos2, podest2026tirex2generalizingtirexmultivariate, das2024timesfm}. These outputs do not determine a joint distribution over future
trajectories. For example, marginal forecasts alone do not specify whether
high values in different channels or at neighboring horizons occur together.
Sampling coordinates independently imposes one particular coupling; it does
not identify dependence from the marginal outputs.

Copulas provide a framework for modeling dependence separately from
marginal distributions. A copula is a multivariate cumulative distribution
function whose univariate marginals are uniform on the unit interval $[0,1]$.
Sklar's theorem states that any joint cumulative distribution function
can be expressed as a copula applied to its marginal cumulative
distribution functions \citep{sklar1959fonctions}.
In our setting, the TSFM supplies the predictive marginals at each
channel and horizon, while the copula specifies how outcomes across
these coordinates co-occur.
A Gaussian copula models this dependence using correlated Gaussian
latent variables, which are transformed into marginally uniform
probability levels and then mapped through the predictive quantile
functions \citep{wen2019quantilecopula}.
Gaussian copula assumes a Gaussian structure only for the latent variables used to model dependence; it does not assume that the TSFM's predictive marginal distributions are Gaussian.

Dependence reconstruction is established in probabilistic forecasting,
including the Schaake Shuffle and ensemble copula coupling
\citep{
TheSchaakeShuffleAMethodforReconstructingSpaceTimeVariabilityinForecastedPrecipitationandTemperatureFields,
Schefzik_2013}.
For multi-step TSFMs, \citet{baron2025correlated} use a temporal Gaussian
copula to generate correlated sample paths from marginal forecasts. Related
approaches have also been explored for pretrained forecasting models
\citep{redhead2026copuadapt,benechehab2025adaptsadaptingunivariatefoundation}.
In complementary theoretical work, \citet{zhang2025fdm} construct
strictly proper scoring rules for continuous Markov processes by averaging
strictly proper scores of two-time joint distributions. Their result
provides a formal example of why joint-time distributions carry identifying
information beyond one-time marginals. It motivates evaluating temporal
dependence separately, but provides no identification or performance
guarantee for our discrete copula baselines or targeted variogram diagnostics.

We use established copula constructions as deliberately simple baselines
and ask: \emph{when frozen TSFM marginals are held fixed, what does
channel--horizon coupling itself contribute?} Our contribution is a
controlled empirical evaluation that holds every finite marginal sample
set exactly fixed. A factorial Gaussian construction and a Schaake-style
historical rank baseline expose temporal and channel coupling effects
separately and show how much these simple choices already improve the
evaluated joint diagnostics. Complementary direct-sampling and inference-mode
experiments assess whether the targeted gains persist across scenario
budgets and when marginal forecasts condition on multivariate history.

\section{Controlled Coupling of Forecast Marginals}
\label{sec:setup}

We instantiate established dependence models using pre-forecast history.
The separable Gaussian construction is an intentionally simple baseline
whose temporal and channel components can be evaluated separately.

\paragraph{Copula representation.}
At a fixed forecast origin, let $F_{d,h}$ be the predictive marginal CDF
constructed from a frozen TSFM's quantile outputs and let
$Q_{d,h}=F_{d,h}^{-1}$ denote its quantile function, for channels
$d=1,\ldots,D$ and horizons $h=1,\ldots,H$.
Conditioning on the observed history up to the forecast origin is suppressed in the notation. A copula $C:[0,1]^{DH}\to[0,1]$ is a joint CDF with uniform marginals
\citep{Schefzik_2013}; it couples the forecast marginals into
\begin{equation}
F_C(y)
=
C\!\left(\bigl(F_{d,h}(y_{d,h})\bigr)_{d,h}\right).
\label{eq:copula-representation}
\end{equation}
Thus, the marginal forecasts and their coupling specify distinct components
of a joint forecast. Quantile construction is detailed in
Appendix~\ref{app:quantiles}.

\paragraph{Gaussian copula.}
For a correlation matrix $R\in\mathbb R^{DH\times DH}$,
the Gaussian copula is
\begin{equation}
C_R(\boldsymbol u)
=
\Phi_R\!\left(\Phi^{-1}(u_1),\ldots,\Phi^{-1}(u_{DH})\right),
\label{eq:gaussian-copula}
\end{equation}
where $\Phi$ is the standard normal CDF and $\Phi_R$ is the CDF
of $\mathcal N(\mathbf 0,R)$
\citep{wen2019quantilecopula,baron2025correlated}.
For the $n$-th sample path, let
$\mathbf z^{(n)}\in\mathbb R^{DH}$ stack the entries of the latent
matrix $Z^{(n)}\in\mathbb R^{D\times H}$ by channel, with
$z^{(n)}_{(d-1)H+h}=Z^{(n)}_{d,h}$.
Thus, horizons are contiguous within each channel.
We generate forecast sample paths by
\begin{equation}
\mathbf z^{(n)}\sim\mathcal N(\mathbf 0,R),
\qquad
U^{(n)}_{d,h}=\Phi\!\left(Z^{(n)}_{d,h}\right),
\qquad
\widehat Y^{(n)}_{d,h}
=
Q_{d,h}\!\left(U^{(n)}_{d,h}\right).
\label{eq:gaussian-direct-sampling}
\end{equation}
Because $R$ has unit diagonal, each $U^{(n)}_{d,h}$ is uniformly
distributed on $(0,1)$, and each $\widehat Y^{(n)}_{d,h}$ has
marginal CDF $F_{d,h}$.
Changing $R$ therefore changes the coupling while preserving
each predictive marginal in distribution.
The matrix $R$ specifies correlations between the latent Gaussian
variables, not generally the Pearson correlations between the
final forecast values.

\paragraph{Exactly fixed empirical marginals.}
Direct draws need not yield identical finite marginal sample sets, even when
the underlying marginal distributions are identical. Our primary experiment
therefore fixes an ordered marginal sample multiset at each coordinate:
\begin{equation}
\mathcal S_{d,h}
=
\left\{Q_{d,h}\!\left(\frac{i-\frac12}{N}\right)\right\}_{i=1}^N,
\qquad
\widehat Y^{(n)}_{d,h}
=
\mathcal S_{d,h}\!\left[\pi_{d,h}(n)\right].
\label{eq:fixed-marginal-coupling}
\end{equation}
Every method receives the same $\mathcal S_{d,h}$ and may change only the
permutations $\pi_{d,h}$. For Gaussian coupling, $\pi_{d,h}(n)$ is the rank
of $Z^{(n)}_{d,h}$ among the $N$ latent draws at that coordinate, in ascending
order. Since $\Phi$ is monotone, these are also the ranks of the copula
probabilities. We use these ranks, rather than the probabilities themselves,
to assign the fixed values. This preserves every empirical marginal exactly
while varying how its values co-occur with those at other coordinates.
All coupling procedures leave the backbone frozen and use only pre-forecast
information.

\paragraph{Factorial Gaussian coupling.}
We estimate $R_{\mathrm{ch}}\in\mathbb R^{D\times D}$ from same-time
correlations of rank-Gaussianized channel histories. For temporal dependence,
we adopt the AR(1)-style form of \citet{baron2025correlated},
$R_{\mathrm{time}}[h,h']=\rho^{|h-h'|}$, where our $\rho$ is the median
channel-wise lag-one correlation of the rank-Gaussianized history.
Including or excluding each relation gives the $2\times2$ factorial
\begin{equation}
\begin{aligned}
R_{\mathrm{IID}}
&=I_D\otimes I_H,
&
R_{\mathrm{Temporal}}
&=I_D\otimes R_{\mathrm{time}},
\\
R_{\mathrm{Channel}}
&=R_{\mathrm{ch}}\otimes I_H,
&
R_{\mathrm{Channel\text{-}Time}}
&=R_{\mathrm{ch}}\otimes R_{\mathrm{time}}.
\end{aligned}
\label{eq:factorial-correlations}
\end{equation}
Each matrix supplies the latent draws used for rank reassignment above.
The separable Channel--Time construction keeps the baseline simple and
supports a controlled decomposition of temporal and channel effects.
It restricts the representable dependence and need not match the true
conditional copula.
Estimation and factorized Gaussian sampling are detailed in
Appendix~\ref{app:couplings}.

\paragraph{Empirical coupling and controls.}
As a non-parametric alternative, a Schaake-style construction
\citep{
TheSchaakeShuffleAMethodforReconstructingSpaceTimeVariabilityinForecastedPrecipitationandTemperatureFields,
Schefzik_2013}
reorders the same fixed samples using historical $D\times H$ rank templates.
We also replace $R_{\mathrm{ch}}$ with an equicorrelation relation or randomly
relabeled historical relations to test the value of correctly aligned,
heterogeneous channel structure. Historical templates and controls are
detailed in Appendices~\ref{app:schaake} and~\ref{app:negative-controls},
respectively.

\paragraph{Direct copula sampling.}
The controlled experiment isolates coupling effects by fixing the
finite marginal sample set at every coordinate.
As a complementary evaluation, we remove this constraint and
generate sample paths directly from each coupling.
Gaussian variants map copula probabilities through the predictive
quantile functions via Equation~\eqref{eq:gaussian-direct-sampling};
historical-template sampling is detailed in
Appendix~\ref{app:direct-sampling}.
Because finite marginal sample sets may differ across methods,
this is an end-to-end evaluation rather than a coupling-only comparison.

\section{Experiments and Results}
\label{sec:experiments}

\paragraph{Experimental setup.}
We evaluate 14 multivariate forecasting tasks from FEV-bench
\citep{shchur2026fevbenchrealisticbenchmarktime}
using three frozen probabilistic backbones: Chronos-2, TimesFM-3,
and TiRex-2
\citep{ansari2025chronos2,das2024timesfm,
podest2026tirex2generalizingtirexmultivariate}.
The primary benchmark uses native multivariate inference and the
fixed-marginal protocol of Section~\ref{sec:setup} with
$N\in\{10,50,100\}$ sample paths.

We report Energy Score and two targeted Variogram Scores:
Ch.-VS evaluates same-horizon cross-channel pairs, while Time-VS
evaluates cross-horizon pairs within a channel.
All scores are lower-is-better and use history-based channel
standardization (Appendix~\ref{app:metrics}).

For dataset $d$ and backbone $b$, let $S_{d,b}(M)$ denote the mean
score of method $M$ on the evaluation set shared by all methods.
We report IID-relative skill,
\begin{equation}
\operatorname{Skill}_{d,b}(M)
=
100\left(
1-\frac{S_{d,b}(M)}{S_{d,b}(\mathrm{IID})}
\right).
\label{eq}
\end{equation}
Skill scores are averaged equally over backbones within each dataset
and then over datasets. Positive values indicate improvement
over IID, while zero indicates equal performance.
Aggregation and uncertainty estimation are detailed in Appendices~\ref{app:skill} and~\ref{app:bootstrap}.

\begin{table}[t]
\centering
\small
\setlength{\tabcolsep}{4pt}
\caption{IID-relative skill (\%) at three scenario budgets. All three backbones use native multivariate inference. Controlled coupling fixes every empirical marginal exactly across methods within a budget; direct sampling permits finite marginal variation. Positive values favor the method over IID, whose skill is zero throughout.}
\label{tab:main}
\begin{tabular}{llrrr@{\hspace{9pt}}rrr}
\toprule
& & \multicolumn{3}{c}{\textbf{Controlled: fixed marginals}} & \multicolumn{3}{c}{\textbf{Direct copula sampling}} \\
\cmidrule(lr){3-5}\cmidrule(lr){6-8}
$N$ & Method & Energy & Ch.-VS & Time-VS & Energy & Ch.-VS & Time-VS \\
\midrule
10 & Temporal & 2.83 & 0.21 & 19.61 & 2.61 & -0.40 & 22.83 \\
 & Channel & 0.02 & 4.23 & -0.02 & -0.02 & 6.65 & -0.17 \\
 & Channel--Time & 2.89 & 4.25 & 19.51 & 2.62 & 6.94 & 22.79 \\
 & Schaake & 2.54 & 4.12 & 17.99 & -1.48 & 6.87 & 21.11 \\
\midrule
50 & Temporal & 2.66 & -0.01 & 21.48 & 2.69 & 0.07 & 21.94 \\
 & Channel & 0.02 & 4.66 & -0.01 & 0.02 & 5.10 & -0.02 \\
 & Channel--Time & 2.55 & 4.63 & 21.47 & 2.56 & 5.22 & 21.94 \\
 & Schaake & 1.76 & 4.15 & 18.93 & 1.02 & 4.29 & 20.05 \\
\midrule
100 & Temporal & 2.65 & 0.00 & 21.64 & 2.64 & 0.01 & 21.87 \\
 & Channel & 0.02 & 4.71 & -0.02 & 0.04 & 4.89 & 0.02 \\
 & Channel--Time & 2.52 & 4.69 & 21.61 & 2.51 & 4.94 & 21.86 \\
 & Schaake & 1.79 & 3.92 & 20.06 & 1.31 & 3.54 & 20.52 \\
\bottomrule
\end{tabular}
\end{table}

\paragraph{Simple baselines improve targeted joint scores.}
With empirical marginals held exactly fixed, Temporal improves Time-VS
and Channel improves Ch.-VS (Table~\ref{tab:main}).
Channel--Time retains both targeted improvements, while Schaake also
improves both diagnostics. These patterns persist across scenario budgets.
The near-zero effects of Temporal on Ch.-VS and Channel on Time-VS
are expected by construction.
Channel's Ch.-VS improvement is accompanied by little change in Energy
Score, indicating that Energy Score is less responsive to the channel-only
intervention in this benchmark.
Additional scores and paired uncertainty are reported in
Appendix~\ref{app:full-results}.

\paragraph{Channel alignment and temporal persistence.}
Historical Channel--Time yields lower aggregate Ch.-VS than the
equicorrelation and randomly relabeled controls.
The paired raw-score comparisons support the value of correctly aligned,
heterogeneous historical channel structure beyond the two tested controls
(Appendix~\ref{app:negative-controls}).

For temporal coupling, a fixed high-persistence coefficient recovers much
of the Time-VS improvement obtained with historical estimation.
The paired comparison does not establish an additional gain from
estimating the temporal coefficient over this fixed baseline.
Thus, substantial temporal gains are already obtained by a simple
persistence assumption (Appendix~\ref{app:temporal-controls}).

\begin{figure}[t]
\centering
\includegraphics[width=0.88\linewidth]{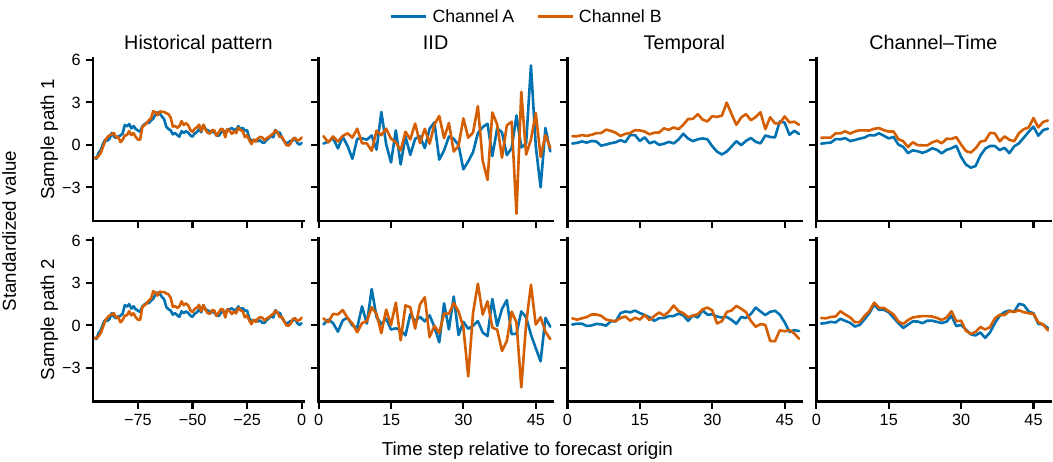}
\caption{
Illustration of direct multivariate sample-path generation using
frozen Chronos-2 marginal forecasts on a fixed synthetic
multivariate history.
Each row shows one prespecified sampled future, with blue and
orange denoting Channels A and B.
The left column repeats the observed pre-forecast history; the
remaining columns compare IID, Temporal, and Channel--Time coupling
using the same coordinate-wise marginal forecasts and common
Gaussian draws.
Temporal coupling introduces within-channel temporal dependence,
whereas Channel--Time additionally incorporates the estimated
historical cross-channel relation.
}
\label{fig:joint-movement}
\end{figure}

\paragraph{Direct sample-path generation.}
We replace fixed-marginal rank reassignment with direct copula
sampling and evaluate $N\in\{10,50,100\}$ over 30 sampling seeds
on a common evaluation set.
The targeted improvements persist across sample sizes.
At $N=100$, Channel--Time achieves $4.94\%$ Ch.-VS skill and
$21.86\%$ Time-VS skill, with lower scores than IID on both
diagnostics in all 30 seeds
(Table~\ref{tab:main}; Appendix~\ref{app:direct-sampling}).

Figure~\ref{fig:joint-movement} illustrates these coupling effects
using frozen TSFM marginal forecasts on a fixed synthetic
multivariate history.
Temporal coupling introduces within-channel temporal structure,
while Channel--Time additionally incorporates the historical
cross-channel relation.

\paragraph{Multivariate conditioning.}
The controlled Channel--Time versus IID comparison yields
positive joint-score skill under both univariate and native
multivariate inference for all three backbones
(Appendix~\ref{app:inference-mode-results}).
Post-hoc coupling therefore remains useful even when the marginal
forecasts already condition on multivariate history.

\paragraph{Direct comparisons and variation across datasets.}
At controlled $N=100$, Channel--Time improves Energy skill over Schaake
by $0.72$ percentage points, while its Energy and global-max CRPS differences from
Temporal remain uncertain (Appendix~\ref{app:paired-couplings}).
Channel--Time has positive Time-VS skill on all 14 datasets and positive
Ch.-VS skill on 11, demonstrating that the aggregate channel gain is
not universal. Dataset-level results are reported in Appendix~\ref{app:dataset-effects}.

\section{Conclusion and Limitations}
\label{sec:conclusion}

Established Gaussian and historical rank couplings provide simple,
training-free baselines for constructing multivariate sample paths from
frozen TSFM marginals. In our controlled benchmark, these baselines
improve targeted joint-forecast diagnostics while preserving every
coordinate-wise empirical marginal sample set. The targeted gains persist
at $N=10,50,100$ under controlled coupling and direct sampling with
native multivariate inference. A fixed persistence coefficient also
retains much of the temporal gain, reinforcing the value of simple
dependence baselines.
The contribution is a controlled empirical reference showing what simple
historical coupling already contributes and against which richer
dependence models can be evaluated.

\paragraph{Limitations.}
Our findings are limited to the evaluated tasks and backbones.
Historical couplings require pre-forecast dependence to remain
informative about future outcomes, which may fail under regime
or dependence shifts.
Historical rank correlations are proxies for conditional future
dependence; the observed diagnostic gains do not establish recovery of
the true conditional copula or improvements in downstream decisions.
Our separable Gaussian copula imposes a horizon-invariant latent
channel relation and a shared temporal correlation structure,
limiting its ability to represent channel-pair-specific lead--lag
patterns and asymmetric or non-Gaussian tail dependence.
Schaake-style coupling depends on sufficient, relevant historical
rank templates.

\paragraph{Future work.}
We plan to continue studying how existing pretrained TSFMs can be used
to generate multivariate forecast trajectories. Motivated by recent
work on trajectory forecasting \citep{yao2026pathfmfastgenerativeforecasting}, we will investigate
lightweight dependence generators that condition on the observed history
and couple frozen TSFM marginals. We will evaluate whether these extensions improve trajectory
quality and sampling efficiency over the historical coupling baselines
studied here.

\clearpage
\bibliographystyle{plainnat}
\bibliography{copula_paper_references}

\clearpage
\appendix
\section{Additional Experimental Details}
\label{app:experimental-details}

\subsection{Benchmark population and backbones}
\label{app:datasets}

We evaluate 14 multivariate forecasting tasks from FEV-bench
\citep{shchur2026fevbenchrealisticbenchmarktime}.
Table~\ref{tab:benchmark-population} reports the target dimensionality $D$,
forecast horizon $H$, observed context length $L$, and number of evaluated
item--origin cells per backbone. One additional declared UCI Air Quality
1H task is excluded upstream because its evaluation windows do not contain
finite ground truth over the complete forecast horizon.

\begin{table}[h]
\centering
\small
\setlength{\tabcolsep}{5pt}
\caption{Core benchmark population. ``Cells'' denotes evaluated item--origin
pairs per backbone.}
\label{tab:benchmark-population}
\begin{tabular}{lrrrr}
\toprule
Dataset & $D$ & $H$ & $L$ & Cells \\
\midrule
UCI Air Quality 1D & 4 & 28 & 249 & 3 \\
ETT 1D & 7 & 28 & 164 & 16 \\
ETT 1H & 7 & 168 & 2048 & 16 \\
ETT 15min & 7 & 96 & 2048 & 16 \\
Jena Weather 1H & 21 & 24 & 2048 & 8 \\
BizITObs L2C 1H & 7 & 24 & 2048 & 8 \\
UK COVID 1D, new & 3 & 28 & 168 & 32 \\
UK COVID 1D, cumulative & 3 & 28 & 168 & 32 \\
UK COVID 1W, new & 3 & 8 & 73 & 16 \\
UK COVID 1W, cumulative & 3 & 8 & 73 & 16 \\
Boomlet & 28 & 60 & 2048 & 8 \\
FRED-MD CEE & 3 & 12 & 558 & 8 \\
FRED-QD CEE & 3 & 8 & 106 & 8 \\
GVAR & 6 & 8 & 98 & 32 \\
\bottomrule
\end{tabular}
\end{table}

We use three frozen probabilistic TSFMs: Chronos-2
\citep{ansari2025chronos2}, TimesFM-3
\citep{das2024timesfm}, and TiRex-2
\citep{podest2026tirex2generalizingtirexmultivariate}.
The primary benchmark uses each backbone's native multivariate target
interface. All backbone parameters remain frozen. Coupling statistics are
estimated from pre-forecast history without gradient-based
training or task-specific fine-tuning.

\subsection{Metrics and history standardization}
\label{app:metrics}

Joint scores are computed after channel-wise standardization using only
observations available before the forecast origin. For channel $d$,
\[
\widetilde y_{d,t}
=
\frac{y_{d,t}-\mu_{d,\mathrm{hist}}}
{\sigma_{d,\mathrm{hist}}+10^{-8}},
\]
where $\mu_{d,\mathrm{hist}}$ and $\sigma_{d,\mathrm{hist}}$ are the mean
and population standard deviation of finite values in the observed context.
The same transformation is applied to forecast samples and realized future
values. No future observation enters these statistics.

For forecast samples $\{\mathbf x^{(n)}\}_{n=1}^{N}$ and realization
$\mathbf y$, the reported Energy Score uses the fair ensemble form
\[
\operatorname{ES}
=
\frac{1}{N}
\sum_{n=1}^{N}
\left\|
\mathbf x^{(n)}-\mathbf y
\right\|_2
-
\frac{1}{N(N-1)}
\sum_{1\le n<m\le N}
\left\|
\mathbf x^{(n)}-\mathbf x^{(m)}
\right\|_2.
\]

For a set of coordinate pairs $\mathcal P$, the empirical Variogram Score
of order $p$ is
\[
\operatorname{VS}_{p,\mathcal P}
=
\sum_{(i,j)\in\mathcal P}
\left(
|y_i-y_j|^p
-
\frac{1}{N}
\sum_{n=1}^{N}
|x_i^{(n)}-x_j^{(n)}|^p
\right)^2.
\]
We use $p=0.5$. The two primary diagnostic pair sets are
\[
\begin{aligned}
\mathcal P_{\mathrm{ch}}
&=
\{((d,h),(d',h)):d<d'\},\\
\mathcal P_{\mathrm{time}}
&=
\{((d,h),(d,h')):h<h'\}.
\end{aligned}
\]
Thus, Channel Variogram Score (Ch.-VS) measures same-horizon
cross-channel structure, whereas Temporal Variogram Score (Time-VS)
measures cross-horizon structure within a channel. If more than 2,000
pairs are available, a fixed method-independent subset of 2,000 pairs is
used for that evaluation cell.

For a scalar predictive ensemble $\{x^{(n)}\}_{n=1}^{N}$ and realization
$y$, we compute
\[
\operatorname{CRPS}
=
\frac{1}{N}\sum_{n=1}^{N}|x^{(n)}-y|
-
\frac{1}{2N^2}
\sum_{n=1}^{N}\sum_{m=1}^{N}
|x^{(n)}-x^{(m)}|.
\]

We additionally evaluate two scalar functionals of each standardized future
field,
\[
G(\mathbf y)
=
\frac{1}{DH}\sum_{d=1}^{D}\sum_{h=1}^{H}y_{d,h},
\qquad
M(\mathbf y)
=
\max_{d,h} y_{d,h},
\]
and report CRPS for the induced predictive distributions of
$G(\mathbf Y)$ and $M(\mathbf Y)$ as Global-mean and Global-max CRPS,
respectively.
\subsection{Dataset-equal aggregation and relative skill}
\label{app:skill}

Absolute score scales differ across tasks because their dimensionalities,
forecast horizons, and pair counts differ. For cross-dataset comparisons we
therefore use IID-relative skill. Let $S_{d,b}(M)$ denote the mean score of
method $M$ on dataset $d$ and backbone $b$, computed on the exact support
shared with IID. We define
\[
\operatorname{Skill}_{d,b}(M)
=
100\left(
1-\frac{S_{d,b}(M)}
{S_{d,b}(\mathrm{IID})}
\right).
\]
Positive values indicate lower scores than IID.

For the overall summary, backbones are averaged equally within each dataset,
and datasets are then weighted equally:
\[
\operatorname{Skill}(M)
=
\frac{1}{|\mathcal D|}
\sum_{d\in\mathcal D}
\left[
\frac{1}{|\mathcal B|}
\sum_{b\in\mathcal B}
\operatorname{Skill}_{d,b}(M)
\right].
\]
The primary all-method comparison is restricted to the 654 evaluation cells
on which IID, Temporal, Channel, Channel--Time, and Schaake are all
available. This support spans all 14 datasets and three backbones.
The cell IDs are unchanged across $N\in\{10,50,100\}$ in the
controlled benchmark and in the direct-sampling comparison. Within a
budget, every controlled method uses the same finite marginal value
set; changing $N$ also changes the quantile grid defining that set.

\subsection{Hierarchical uncertainty estimation}
\label{app:bootstrap}

All reported bootstrap intervals use 10,000 hierarchical percentile
replicates after sampling seeds are averaged within a cell. The original
method-versus-IID intervals retained in Appendix~\ref{app:main-paired}
and channel-control intervals in Appendix~\ref{app:negative-controls}
resample datasets, backbone-specific items, and origins. Their statistic
is a nested origin--item--dataset mean, which can differ from the
plain-cell aggregation used for headline scores when items have unequal
origin counts. Those intervals are identified as intervals for the nested
raw-score difference; they are not intervals for the main-table skill.

The direct Channel--Time-versus-Temporal/Schaake comparisons instead
resample the same underlying item--origin units jointly across backbones and recompute the exact
headline cell-mean and skill statistics within every draw. The primary
normalized contrasts are reported in Appendix~\ref{app:paired-couplings}. The temporal-control
analysis retains its own nested bootstrap convention, detailed in
Appendix~\ref{app:temporal-controls}. These procedures and their target
statistics are distinguished rather than treating their intervals as
interchangeable.

\section{Coupling Details}
\label{app:couplings}

\subsection{Marginal quantile construction}
\label{app:quantiles}

All three backbones provide predictive quantiles at levels
$\{0.1,0.2,\ldots,0.9\}$.
Within each channel--horizon coordinate, we first monotonically
rearrange the predicted quantile knots to remove crossings and then
use piecewise-linear interpolation between adjacent interior knots.

For a fixed coordinate, let $q_\tau$ denote the rearranged quantile
at level $\tau$.
The tails use logarithmic quantile extrapolation, with scales
determined by the adjacent outer quantile gaps.
For $u<0.1$,
\[
Q(u)
=
q_{0.1}
-
(q_{0.2}-q_{0.1})
\log_2\!\left(\frac{0.1}{u}\right),
\]
and for $u>0.9$,
\[
Q(u)
=
q_{0.9}
+
(q_{0.9}-q_{0.8})
\log_2\!\left(\frac{0.1}{1-u}\right).
\]
The same inverse-quantile construction is used by every coupling
method.

\subsection{Historical channel relation}
\label{app:channel-relation}

Let $Y_{d,t}$ denote the observed pre-forecast history of channel
$d$.
Each channel is transformed independently to Gaussian scores using
empirical ranks.
If $r_{d,t}$ is the average rank of a finite observation among the
$T_d$ finite observations of channel $d$, we define
\[
u_{d,t}
=
\frac{r_{d,t}}{T_d+1},
\qquad
z_{d,t}
=
\Phi^{-1}(u_{d,t}).
\]
The historical channel relation $R_{\mathrm{ch}}$ is the Pearson
correlation matrix of the rank-Gaussianized channels, computed over
timestamps at which all channels are finite.

If fewer than 32 complete timestamps are available, or if $D<2$,
the implementation falls back to $I_D$; this fallback does not occur
in the Core benchmark.
For numerical stability, we clip eigenvalues at $10^{-6}$, reconstruct
the matrix, rescale it to unit diagonal, and then compute its
Cholesky factor.
No shrinkage estimator is used.

\subsection{Historical temporal relation}
\label{app:temporal-relation}

For each channel, we compute the lag-one Pearson correlation between
consecutive rank-Gaussianized observations, using only pairs for which
both endpoints are finite.
Channels with fewer than two usable pairs or zero variance are
excluded.
The temporal coefficient is the median across usable channels,
clipped to $[-0.99,0.99]$:
\[
R_{\mathrm{time}}[h,h']
=
\rho^{|h-h'|}.
\]

\subsection{Gaussian rank coupling}
\label{app:gaussian-rank-coupling}

For each sampling seed, the Gaussian variants share a common
base-normal tensor
\[
E\in\mathbb R^{N\times D\times H},
\]
and construct
\[
Z^{(n)}
=
L_{\mathrm{ch}}
E^{(n)}
L_{\mathrm{time}}^\top,
\]
where $L_{\mathrm{ch}}L_{\mathrm{ch}}^\top=R_{\mathrm{ch}}$ and
$L_{\mathrm{time}}L_{\mathrm{time}}^\top=R_{\mathrm{time}}$.
The factorial variants replace either factor with the corresponding
identity matrix.

In the controlled fixed-marginal experiment, $\Phi(Z)$ is not
mapped through the inverse quantile function.
Instead, only the rank ordering of $Z_{:,d,h}$ across the $N$
sample paths is used.
The $r$-th smallest fixed marginal value is assigned to the path
whose latent Gaussian value has rank $r$.
Thus, Gaussian coupling only permutes the common finite marginal
samples and introduces no new forecast values.

Using one-based channel and horizon indices, the flattened
channel-major coordinate index is
\[
i=(d-1)H+h.
\]
Under this ordering, the implied full latent Gaussian correlation is
\[
R_{\mathrm{ch}}\otimes R_{\mathrm{time}}.
\]

\subsection{Historical empirical rank coupling}
\label{app:schaake}

As a nonparametric alternative, we use a Schaake-style historical
rank reconstruction
\citep{
TheSchaakeShuffleAMethodforReconstructingSpaceTimeVariabilityinForecastedPrecipitationandTemperatureFields,
Schefzik_2013}.

Candidate $D\times H$ historical blocks are constructed strictly
from observations preceding the forecast origin.
The procedure first steps backward through history in increments of
$H$, preferring the most recent non-overlapping fully finite blocks.
If fewer than $N$ such blocks are available, the spacing is reduced
to one time step and the most recent fully finite overlapping blocks
are used instead. If fewer than $N$ remain, available blocks are
reused by deterministic cycling. Availability is determined at $N=10$:
an evaluation cell needs at least five eligible blocks after the
spacing fallback, and that decision is kept for all three budgets.

In the controlled experiment, the selected blocks provide only rank
templates.
At each coordinate $(d,h)$, their historical values define an
ordering across templates, and the common fixed forecast samples are
permuted to match this ordering.
Hence the empirical marginal sample multiset at every coordinate is
unchanged.

The direct-sampling version and its finite-template behavior are
described separately in Appendix~\ref{app:direct-sampling}.

\subsection{Fixed-marginal audit}
\label{app:fixed-marginal-audit}

We audited the controlled implementation at $N=10,50,100$ over all
stored Core cells and coupling variants. Direct regeneration of the four
Gaussian methods at all 100 sampling seeds and Schaake once per available
cell yields exactly the fixed sorted values in every case, with zero
multiset violations.

This exact finite-marginal identity applies only to the controlled
fixed-marginal experiments.
It does not apply to the direct-sampling experiment in
Appendix~\ref{app:direct-sampling}.

\section{Full Controlled Benchmark Results}
\label{app:full-results}

We report the controlled $N\in\{10,50,100\}$ benchmark on the same
654 evaluation cells
shared by IID, Temporal, Channel, Channel--Time, and Schaake.
All three backbones use native multivariate inference. Within each
cell, the methods receive exactly the same coordinate-wise empirical
marginal sample multisets and differ only in their coupling.
Metrics and aggregation follow Appendices~\ref{app:metrics}
and~\ref{app:skill}.

\subsection{Absolute scores}
\label{app:absolute-results}

Table~\ref{tab:all-method-common} reports dataset-equal absolute scores.
Mean CRPS and Max CRPS evaluate the global mean and global maximum
of the standardized future field, respectively; they are distinct
from coordinate-wise marginal CRPS.

\begin{table}[htbp]
\centering
\small
\setlength{\tabcolsep}{4pt}
\caption{Dataset-equal absolute scores; lower is better. All three backbones use native multivariate inference. Mean and Max denote CRPS of the global mean and global maximum of the standardized future field.}
\label{tab:all-method-common}
\begin{tabular}{llrrrrr}
\toprule
$N$ & Method & Energy & Ch.-VS & Time-VS & Mean CRPS & Max CRPS \\
\midrule
10 & IID & 8.910 & 121.082 & 217.623 & 0.232 & 1.551 \\
 & Temporal & 8.830 & 121.062 & 198.899 & 0.206 & 1.346 \\
 & Channel & 8.906 & 112.819 & 217.602 & 0.225 & 1.537 \\
 & Channel--Time & 8.805 & 112.781 & 199.096 & 0.201 & 1.341 \\
 & Schaake & 8.801 & 111.717 & 202.239 & 0.210 & 1.387 \\
\midrule
50 & IID & 9.209 & 119.941 & 216.091 & 0.229 & 1.620 \\
 & Temporal & 9.125 & 119.936 & 194.605 & 0.200 & 1.332 \\
 & Channel & 9.204 & 110.727 & 216.115 & 0.220 & 1.571 \\
 & Channel--Time & 9.103 & 110.743 & 194.609 & 0.196 & 1.296 \\
 & Schaake & 9.189 & 111.511 & 200.212 & 0.206 & 1.385 \\
\midrule
100 & IID & 9.250 & 119.842 & 215.973 & 0.229 & 1.643 \\
 & Temporal & 9.164 & 119.839 & 194.150 & 0.199 & 1.335 \\
 & Channel & 9.244 & 110.511 & 215.992 & 0.220 & 1.580 \\
 & Channel--Time & 9.142 & 110.523 & 194.199 & 0.195 & 1.290 \\
 & Schaake & 9.192 & 111.631 & 197.210 & 0.201 & 1.331 \\
\bottomrule
\end{tabular}
\end{table}

\subsection{IID-relative skill}
\label{app:controlled-skill}

Table~\ref{tab:controlled-skill} reports IID-relative skill using the
same support. Ratios are computed within each dataset--backbone pair
before equal averaging over backbones within datasets and then over
datasets. Consequently, these skills cannot be recovered by taking
ratios of the aggregate absolute scores in
Table~\ref{tab:all-method-common}.

\begin{table}[htbp]
\centering
\small
\setlength{\tabcolsep}{4pt}
\caption{IID-relative skill (\%); positive values favor the method over IID. Ratios are formed within dataset--backbone pairs before equal aggregation. All three backbones use native multivariate inference. Exactly the same 654 cells are used at $N=10,50,100$, with empirical marginals fixed across methods within each budget. Mean and Max denote CRPS of the global mean and global maximum of the standardized future field.}
\label{tab:controlled-skill}
\begin{tabular}{llrrrrr}
\toprule
$N$ & Method & Energy & Ch.-VS & Time-VS & Mean CRPS & Max CRPS \\
\midrule
10 & Temporal & 2.83 & 0.21 & 19.61 & 9.44 & 16.84 \\
 & Channel & 0.02 & 4.23 & -0.02 & 2.11 & 1.65 \\
 & Channel--Time & 2.89 & 4.25 & 19.51 & 6.17 & 16.72 \\
 & Schaake & 2.54 & 4.12 & 17.99 & 5.99 & 13.29 \\
\midrule
50 & Temporal & 2.66 & -0.01 & 21.48 & 10.19 & 23.49 \\
 & Channel & 0.02 & 4.66 & -0.01 & 2.47 & 3.96 \\
 & Channel--Time & 2.55 & 4.63 & 21.47 & 6.46 & 24.54 \\
 & Schaake & 1.76 & 4.15 & 18.93 & 4.13 & 18.56 \\
\midrule
100 & Temporal & 2.65 & 0.00 & 21.64 & 10.32 & 24.62 \\
 & Channel & 0.02 & 4.71 & -0.02 & 2.53 & 4.72 \\
 & Channel--Time & 2.52 & 4.69 & 21.61 & 6.56 & 26.07 \\
 & Schaake & 1.79 & 3.92 & 20.06 & 5.83 & 21.34 \\
\bottomrule
\end{tabular}
\end{table}

\subsection{Paired uncertainty for the controlled factorial}
\label{app:main-paired}

For the original $N=10$ Gaussian factorial, we retain the reported
method-versus-IID uncertainty analysis on the same 654-cell support:
\[
\Delta(M)=S(M)-S(\mathrm{IID}),
\]
where $S(M)$ denotes the dataset-equal mean raw score. Negative
$\Delta$ favors the coupling method. Its accompanying legacy bootstrap
uses a nested origin--item--dataset mean rather than exactly this
cell-mean statistic (Appendix~\ref{app:bootstrap}). Table~\ref{tab:main-paired-uncertainty}
therefore prints the nested point estimate with its own 95\% interval.

Table~\ref{tab:main-paired-uncertainty} repeats the corresponding
headline skills alongside these intervals. The intervals target the nested
raw-score difference, not the skill percentages, which use
dataset--backbone-specific normalization. Direct comparisons between
coupling methods using the exact headline aggregation are reported in
Appendix~\ref{app:paired-couplings}.

\begin{table}[htbp]
\centering
\small
\setlength{\tabcolsep}{5pt}
\caption{
Headline IID-relative skill and the original nested paired raw-score
uncertainty at $N=10$ on 654 cells. The nested difference averages
origins within backbone-specific items and then datasets; negative
values favor the coupling method. Brackets give 95\% intervals for
that nested statistic, not for the headline skill.
}
\label{tab:main-paired-uncertainty}
\begin{tabular}{llrr}
\toprule
Method & Metric & Skill (\%) & Nested $\Delta$ [95\% CI] \\
\midrule
\multirow{3}{*}{Temporal}
& Energy  & 2.83  & $-0.0797\;[-0.135,\,-0.040]$ \\
& Ch.-VS  & 0.21  & $-0.0199\;[-0.107,\,0.073]$ \\
& Time-VS & 19.61 & $-18.74\;[-27.7,\,-10.8]$ \\
\addlinespace
\multirow{3}{*}{Channel}
& Energy  & 0.02  & $-0.0037\;[-0.007,\,-0.001]$ \\
& Ch.-VS  & 4.23  & $-8.264\;[-15.9,\,-1.93]$ \\
& Time-VS & -0.02 & $-0.0212\;[-0.169,\,0.137]$ \\
\addlinespace
\multirow{3}{*}{Channel--Time}
& Energy  & 2.89  & $-0.1048\;[-0.178,\,-0.054]$ \\
& Ch.-VS  & 4.25  & $-8.302\;[-15.9,\,-1.98]$ \\
& Time-VS & 19.51 & $-18.54\;[-27.4,\,-10.7]$ \\
\bottomrule
\end{tabular}
\end{table}

\subsection{Factorial interpretation}
\label{app:factorial-results}

\paragraph{Construction-implied invariances.}
For a same-horizon cross-channel pair $((d,h),(d',h))$, Temporal
coupling has zero latent correlation, as does IID. Its bivariate law
relevant to Ch.-VS is therefore unchanged. Conversely, Channel
coupling leaves the within-channel cross-horizon bivariate laws
relevant to Time-VS unchanged from IID. The small observed differences
on these untargeted axes are sampling variation; their paired
intervals include zero. These contrasts are design checks, not
independent discoveries about forecasting ability.

\paragraph{Targeted improvements.}
The empirical result is that the historical relations improve their
targeted diagnostics. Temporal lowers Time-VS by $18.72$ raw-score
units relative to IID, and Channel lowers Ch.-VS by $8.263$ at $N=10$.
The corresponding nested raw-score intervals exclude zero.
Channel--Time also improves both
diagnostics relative to IID. Schaake has positive skill on both axes
in Table~\ref{tab:controlled-skill}, although its paired uncertainty
is not included in Table~\ref{tab:main-paired-uncertainty}.

\paragraph{Metric response and comparison scope.}
Channel achieves $4.23\%$ Ch.-VS skill but only $0.02\%$ Energy
skill. Thus, Energy Score is less responsive to this channel-only
intervention under the reported relative-skill measure; its small
paired Energy interval below zero does not imply a large effect.
The intervals in Table~\ref{tab:main-paired-uncertainty} compare each
method with IID. In particular, the Energy skills of Temporal
($2.83\%$) and Channel--Time ($2.89\%$) do not establish that
Channel--Time outperforms Temporal. The primary direct comparisons in
Appendix~\ref{app:paired-couplings} address this separate question.

\section{Jointization skill by inference mode}
\label{app:inference-mode-results}

Native multivariate inference can condition each coordinate-wise marginal
forecast on other channels, but this does not by itself specify a joint
distribution over future outcomes. We therefore evaluate whether post-hoc
dependence reconstruction remains useful under both univariate and native
multivariate backbone inference.

Within each backbone and inference mode, IID and Channel--Time use exactly
the same coordinate-wise empirical marginal samples under the controlled
$N=10$ protocol. Thus, each comparison changes only the coupling while
holding the mode-specific marginal forecasts fixed.

Table~\ref{tab:inference-mode-skill} reports IID-relative skill,
\[
\operatorname{Skill}_{d,b}(M)
=
100\left(
1-\frac{S_{d,b}(M)}
{S_{d,b}(\mathrm{IID})}
\right),
\]
where the ratio is computed within each dataset $d$ for a fixed backbone
$b$, and then averaged equally across datasets. Positive values indicate
lower scores than IID.

These results use the full 657-cell Gaussian support (219 cells per
backbone), on which IID and Channel--Time are both available. This differs
from the 654-cell common support used by the all-method comparison in
Table~\ref{tab:main}, which additionally requires the
Schaake-style coupling to be available. Consequently, the three Multi rows
below should not be averaged and expected to reproduce the overall
Channel--Time values in Table~\ref{tab:main}.

\begin{table}[h]
\centering
\small
\setlength{\tabcolsep}{4.5pt}
\caption{
IID-relative skill (\%) of Channel--Time coupling under univariate
and native multivariate backbone inference, using the controlled $N=10$
fixed-marginal protocol.
Within each backbone--mode pair, IID and Channel--Time use identical
empirical marginal samples. Skill is computed within each dataset before
cross-dataset averaging. Mean and Max denote CRPS of the global mean and global
maximum of the standardized future field.
}
\label{tab:inference-mode-skill}
\begin{tabular}{llrrrrr}
\toprule
Backbone & Mode
& Energy
& Ch.-VS
& Time-VS
& Mean CRPS
& Max CRPS \\
\midrule

Chronos-2
& Uni
& 2.33
& 2.84
& 14.45
& 13.77
& 13.47 \\
& Multi
& 2.45
& 3.24
& 17.14
& 13.22
& 17.09 \\

\addlinespace

TimesFM-3
& Uni
& 3.48
& 3.35
& 20.91
& 11.24
& 21.13 \\
& Multi
& 3.72
& 4.42
& 22.18
& 13.54
& 17.35 \\

\addlinespace

TiRex-2
& Uni
& 2.25
& 2.07
& 17.76
& 12.37
& 13.67 \\
& Multi
& 2.47
& 3.15
& 18.90
& 12.55
& 13.88 \\

\bottomrule
\end{tabular}
\end{table}

Channel--Time jointization yields positive IID-relative skill for every
backbone and inference mode shown. In particular, under native multivariate
inference, Ch.-VS skill ranges from $3.15\%$ to $4.42\%$ and Time-VS skill
from $17.14\%$ to $22.18\%$. Thus, allowing the backbone to condition its
marginal forecasts on cross-channel history does not eliminate the additional
benefit of coupling future outcomes.

We do not interpret the difference between the Uni and Multi rows as the
causal effect of multivariate inference. Changing the inference mode also
changes the frozen marginal forecasts themselves, and the magnitude of the
jointization gain is not uniformly larger under Multi across all metrics and
backbones. The relevant conclusion is instead that the IID-to-Channel--Time
gain remains present within each inference mode.
\section{Negative Controls for Cross-Channel Structure}
\label{app:negative-controls}

The controlled factorial compares historical channel coupling with
independence. Here we test a narrower question: whether the historical
channel relation provides information beyond its average signed
correlation or a randomly relabeled correlation structure. Both
controls retain the same temporal relation and fixed empirical
marginals as historical Channel--Time.

\subsection{Control definitions}
\label{app:negative-control-definitions}

\paragraph{Equicorrelation.}
Let $\bar\rho$ be the mean \emph{signed} off-diagonal entry of the
PSD-repaired historical $R_{\mathrm{ch}}$. We construct
\[
R_{\mathrm{equi}}
=
(1-\bar\rho)I_D+\bar\rho\,\mathbf 1\mathbf 1^\top.
\]
If necessary, $\bar\rho$ is clipped to the valid equicorrelation
interval $(-1/(D-1),1)$ with a $10^{-6}$ numerical margin. Up to
this numerical clipping, the control preserves average signed
correlation while removing pair-specific heterogeneity.

\paragraph{Randomly relabeled historical relation.}
For a random permutation matrix $P$, we construct
\[
R_{\mathrm{perm}}=P R_{\mathrm{ch}}P^\top.
\]
This preserves the multiset of correlation entries and the spectrum
of $R_{\mathrm{ch}}$, but changes their alignment with the observed
channel labels. Results are averaged over five generated permutations
per evaluation cell. Individual permutations can leave some channel
pairs unchanged, especially when $D$ is small. The control therefore
tests random relabeling rather than complete removal of channel-pair
identity.

In each control, the replacement channel matrix is combined with the
same $R_{\mathrm{time}}$ used by historical Channel--Time. Coupling
uses the controlled $N=10$ rank-reassignment protocol, so every
coordinate-wise empirical marginal remains fixed.

\subsection{Paired evaluation and results}
\label{app:negative-control-results}

Table~\ref{tab:negative-controls} reports mean paired raw-score
differences, with historical Channel--Time minus the corresponding
control. Negative differences favor the historical relation.
The displayed differences use plain cell means within each
dataset--backbone pair. The retained intervals use the legacy nested
origin--item--dataset statistic described in Appendix~\ref{app:bootstrap},
whose weighting can differ from those cell means. W/T/L counts
dataset-level wins, ties, and losses for the historical relation.

\begin{table}[htbp]
\centering
\small
\setlength{\tabcolsep}{6pt}
\caption{Historical Channel--Time minus each cross-channel negative control.
Negative differences favor the historical relation. The displayed
$\Delta$ uses cell means, while the 95\% intervals target the nested
raw-score contrast described in Appendix~\ref{app:bootstrap}.
W/T/L counts dataset-level wins, ties, and losses.}
\label{tab:negative-controls}
\begin{tabular}{llrll}
\toprule
Control & Metric & $\Delta$ & Nested 95\% CI & W/T/L \\
\midrule
\multirow{5}{*}{Equicorr.}
& Energy    & $-0.014$ & $[-0.038,\,-0.001]$ & 9/0/5 \\
& Ch.-VS    & $-6.783$ & $[-12.775,\,-1.706]$ & 12/0/2 \\
& Time-VS   & $0.119$  & $[-0.066,\,0.366]$ & 6/0/8 \\
& Mean CRPS & $0.000$  & $[0.000,\,0.001]$ & 5/0/9 \\
& Max CRPS  & $-0.001$ & $[-0.019,\,0.016]$ & 6/1/7 \\
\midrule
\multirow{5}{*}{Permuted}
& Energy    & $-0.025$ & $[-0.053,\,-0.008]$ & 11/0/3 \\
& Ch.-VS    & $-7.728$ & $[-14.105,\,-2.253]$ & 12/0/2 \\
& Time-VS   & $0.042$  & $[-0.106,\,0.230]$ & 5/0/9 \\
& Mean CRPS & $0.000$  & $[-0.001,\,0.001]$ & 6/0/8 \\
& Max CRPS  & $-0.003$ & $[-0.020,\,0.013]$ & 10/1/3 \\
\bottomrule
\end{tabular}
\end{table}

\subsection{Interpretation}

Historical Channel--Time yields lower Ch.-VS than each control on
12 of 14 datasets. Relative to equicorrelation, its mean paired
difference is $-6.78$; the corresponding nested 95\% interval is
$[-12.77,-1.71]$. Relative to random relabeling, the cell-mean
difference is $-7.73$ and the nested interval is $[-14.10,-2.25]$.
These results support the value of
correctly aligned, heterogeneous historical channel structure beyond
the two tested alternatives in this benchmark.

The Energy contrasts also favor the historical relation, with paired
intervals below zero. We do not compare the raw magnitudes of Energy
and Ch.-VS changes because the metrics have different scales.
The supplementary functional CRPS intervals include zero for both
controls.

Time-VS differences are near zero, with both paired intervals
including zero. This is expected because all three conditions share
$R_{\mathrm{time}}$ and differ only in the channel relation.
The within-channel temporal bivariate laws relevant to Time-VS are
therefore unchanged. These temporal contrasts are design checks;
the Ch.-VS contrasts test the information in channel heterogeneity
and alignment.

\section{End-to-End Direct Sampling}
\label{app:direct-sampling}

We evaluate direct copula sampling for each coupling,
reusing the frozen marginal forecasts from the controlled
benchmark without additional backbone inference. Unlike fixed-marginal rank
reassignment, direct generation does not force the realized
coordinate-wise sample multisets to coincide. This experiment is
therefore an end-to-end robustness check, not a coupling-only
comparison with identical empirical marginals.

All scenario budgets $N\in\{10,50,100\}$ use 30 sampling seeds. Scores use the history-based
standardization and definitions in Appendix~\ref{app:metrics}.

\subsection{Direct Gaussian scenario generation}
\label{app:direct-sampling-setup}

For each scenario $n$, we draw
\[
E^{(n)}_{d,h}\overset{\mathrm{iid}}{\sim}\mathcal N(0,1),
\qquad
Z^{(n)}=L_{\mathrm{ch}}E^{(n)}L_{\mathrm{time}}^\top,
\]
and map the latent Gaussian values through the normal CDF and the
frozen marginal quantile functions:
\[
U^{(n)}=\Phi(Z^{(n)}),
\qquad
\widehat Y^{(n)}_{d,h}=Q_{d,h}\!\left(U^{(n)}_{d,h}\right).
\]
The IID variant replaces both factors with identities, Temporal
replaces only the channel factor, and Channel replaces only the
temporal factor. Channel--Time uses both historical factors.
The inverse-quantile construction is unchanged from
Appendix~\ref{app:quantiles}.

\subsection{Schaake direct sampling}
\label{app:direct-sampling-schaake}

The direct Schaake generator forms a finite pool of $K$ distinct
historical rank templates. Candidate blocks are the most recent
non-overlapping $D\times H$ blocks; if fewer than $N$ fully finite
blocks are available, the spacing drops to one time step. All blocks
precede the forecast origin, and repeated copies of a block are removed.
At each coordinate $(d,h)$, the $K$ blocks receive zero-based ranks
$r_{k,d,h}$, with ties broken by their stable block order. The
pseudo-observations are
\[
U_{k,d,h}=\frac{r_{k,d,h}+1}{K+1}.
\]
Each scenario independently draws one template index uniformly with
replacement and maps its pseudo-observations through the frozen
marginal quantile functions, without within-rank-bin jitter.

The number of generated scenarios $N$ can exceed the number of
distinct templates $K$. Some evaluation cells retain only a small
template pool even at $N=100$, so multiple draws can reuse the same
historical rank structure. The pool is constructed separately for
each $N$, and its size can increase with $N$ up to the number of
available historical blocks. This finite-template constraint is relevant when
interpreting scenario-budget sensitivity. We treat Schaake as a
nonparametric historical-copula baseline and do not claim uniform
dominance of Gaussian Channel--Time over it.

\subsection{Sampling-budget sensitivity}
\label{app:direct-sampling-results}

Table~\ref{tab:direct-sampling-all} reports the three joint-score
diagnostics and two supplementary functional CRPS measures.
For dataset $d$, backbone $b$, and method $M$, let $S_{d,b}(M)$ be
the raw score averaged over the 30 sampling seeds and then over the
common evaluation cells of that dataset--backbone pair.
IID-relative skill is
\[
\operatorname{Skill}_{d,b}(M)
=
100\left(1-\frac{S_{d,b}(M)}{S_{d,b}(\mathrm{IID})}\right),
\]
with equal averaging over backbones within each dataset and then
over datasets. Thus, raw scores are averaged over seeds before
forming the ratios. Positive values indicate improvement over IID.

\begin{table}[htbp]
\centering
\small
\setlength{\tabcolsep}{4pt}
\caption{
IID-relative skill (\%) under direct copula sampling.
Values are computed from raw scores averaged over 30 sampling seeds
on the same 654-cell common support. Ratios are computed within dataset--backbone pairs before
dataset-equal aggregation. Positive values indicate improvement over
IID. Mean and Max denote the supplementary functional CRPS measures.
}
\label{tab:direct-sampling-all}
\begin{tabular}{llrrrrr}
\toprule
Method & $N$ & Energy & Ch.-VS & Time-VS & Mean CRPS & Max CRPS \\
\midrule
Temporal
& 10  & 2.61 & -0.40 & 22.83 & 6.15 & 25.06 \\
& 50  & 2.69 & 0.07  & 21.94 & 9.81 & 26.43 \\
& 100 & 2.64 & 0.01  & 21.87 & 9.99 & 26.41 \\
\addlinespace
Channel
& 10  & -0.02 & 6.65 & -0.17 & 1.44 & 5.79 \\
& 50  & 0.02  & 5.10 & -0.02 & 2.26 & 5.86 \\
& 100 & 0.04  & 4.89 & 0.02  & 2.40 & 6.03 \\
\addlinespace
Channel--Time
& 10  & 2.62 & 6.94 & 22.79 & -0.34 & 26.29 \\
& 50  & 2.56 & 5.22 & 21.94 & 5.12  & 28.03 \\
& 100 & 2.51 & 4.94 & 21.86 & 5.78  & 28.21 \\
\addlinespace
Schaake
& 10  & -1.48 & 6.87 & 21.11 & 1.45 & 20.58 \\
& 50  & 1.02  & 4.29 & 20.05 & 3.77 & 22.80 \\
& 100 & 1.31  & 3.54 & 20.52 & 6.33 & 23.45 \\
\bottomrule
\end{tabular}
\end{table}

The targeted Gaussian pattern persists across budgets: Temporal
improves Time-VS, Channel improves Ch.-VS, and Channel--Time
improves both. The untargeted contrasts remain close to zero.
Effect sizes are not identical across budgets; for example,
Channel--Time Ch.-VS skill decreases from $6.94\%$ at $N=10$
to $4.94\%$ at $N=100$, while its Time-VS skill remains above
$21\%$. The conclusion concerns persistence of the targeted
improvements, not invariance of their magnitude.

Temporal and Channel--Time retain approximately $2.5$--$2.7\%$
Energy skill across budgets, whereas Channel-only Energy skill
remains close to zero. Schaake has positive skill on both targeted
VS diagnostics at all three budgets, but its Energy skill changes
from $-1.48\%$ at $N=10$ to $1.31\%$ at $N=100$.

Functional CRPS evaluates distributions of quantities derived from
complete trajectories. Temporal and Channel--Time show positive
global-max CRPS skill across budgets, whereas Channel--Time
global-mean CRPS skill is negative at $N=10$ and positive at the
larger budgets. Small dataset-specific IID baselines can make these
relative ratios sensitive to aggregation. We therefore retain the
functional CRPS measures as supplementary diagnostics rather than
headline results.

\subsection{Coordinate-wise marginal CRPS}
\label{app:direct-sampling-marginal-crps}

Because direct sampling does not fix the finite marginal sample
sets, we separately evaluate coordinate-wise marginal CRPS.
Unlike global-mean and global-max CRPS, this diagnostic evaluates
the individual channel--horizon marginals rather than a functional
of the complete future field.

\begin{table}[htbp]
\centering
\small
\caption{
IID-relative coordinate-wise marginal CRPS skill (\%) under direct
sampling, using the same dataset--backbone normalization and
aggregation as the joint metrics. Positive values favor the method
over IID. These are aggregate marginal-score comparisons, not tests
of equality of finite empirical marginals.
}
\label{tab:direct-marginal-crps}
\begin{tabular}{lrrr}
\toprule
Method & $N=10$ & $N=50$ & $N=100$ \\
\midrule
Temporal      & -0.289 & -0.021 & -0.004 \\
Channel       & -0.153 & 0.010  & 0.015 \\
Channel--Time & -0.381 & -0.009 & -0.013 \\
Schaake       & 0.879  & 0.030  & 0.019 \\
\bottomrule
\end{tabular}
\end{table}

At $N=100$, the magnitude of aggregate marginal-CRPS skill is below
$0.02\%$ for every method, while targeted joint-score gains remain
substantial. Thus, the joint improvements are accompanied by small
aggregate marginal-score differences. This does not establish
equality of the realized empirical marginals or rule out
differences at individual coordinates or evaluation cells.

\subsection{Absolute scores}
\label{app:direct-sampling-absolute}

Table~\ref{tab:direct-sampling-absolute} reports dataset-equal
absolute scores at each budget. Their cross-dataset ratios should
not be used to reconstruct Table~\ref{tab:direct-sampling-all}:
IID-relative skills are computed within dataset--backbone pairs
before aggregation.

\begin{table}[htbp]
\centering
\small
\setlength{\tabcolsep}{3.8pt}
\caption{
Dataset-equal absolute scores under direct stochastic scenario
generation on the 654-cell common support. All metrics are
lower-is-better. Mean and Max denote CRPS of the global mean and
global maximum of the standardized future field.
}
\label{tab:direct-sampling-absolute}
\begin{tabular}{llrrrrr}
\toprule
Method & $N$ & Energy & Ch.-VS & Time-VS & Mean CRPS & Max CRPS \\
\midrule
IID
& 10  & 9.284 & 129.037 & 231.426 & 0.2319 & 1.713 \\
& 50  & 9.291 & 121.577 & 218.964 & 0.2287 & 1.671 \\
& 100 & 9.291 & 120.653 & 217.385 & 0.2283 & 1.666 \\
\addlinespace
Temporal
& 10  & 9.186 & 128.935 & 206.511 & 0.2096 & 1.372 \\
& 50  & 9.200 & 121.608 & 196.400 & 0.2003 & 1.332 \\
& 100 & 9.200 & 120.601 & 194.989 & 0.1994 & 1.327 \\
\addlinespace
Channel
& 10  & 9.282 & 118.271 & 231.519 & 0.2244 & 1.630 \\
& 50  & 9.284 & 111.933 & 218.993 & 0.2200 & 1.591 \\
& 100 & 9.284 & 111.116 & 217.368 & 0.2194 & 1.584 \\
\addlinespace
Channel--Time
& 10  & 9.162 & 118.107 & 206.937 & 0.2103 & 1.325 \\
& 50  & 9.177 & 112.006 & 196.521 & 0.1971 & 1.283 \\
& 100 & 9.178 & 111.107 & 195.099 & 0.1958 & 1.276 \\
\addlinespace
Schaake
& 10  & 9.597 & 118.380 & 209.681 & 0.2204 & 1.439 \\
& 50  & 9.343 & 112.879 & 200.322 & 0.2087 & 1.379 \\
& 100 & 9.271 & 112.223 & 197.370 & 0.2018 & 1.331 \\
\bottomrule
\end{tabular}
\end{table}

\subsection{Synthetic illustration}
\label{app:synthetic-illustration}

Figure~\ref{fig:joint-movement} provides a synthetic illustration
separate from the FEV-bench evaluation.
For channels $d=0,1,2$, we generate
\[
y_{d,t}
=
o_d+s_d\left(
a f_t+\sqrt{1-a^2}\,e_{d,t}
\right),
\]
where $a=0.95$, $\mathbf{o}=(0,10,50)$, and
$\mathbf{s}=(1,2,5)$.
The common factor $f_t$ and channel-specific processes $e_{d,t}$
are mutually independent stationary unit-variance Gaussian AR(1)
processes, with coefficients $0.98$ and $0.90$, respectively.
We use $D=3$, a history length of $L=512$, and a forecast horizon
of $H=48$.

The synthetic realization is fixed at index $0$ with seed
$20260917$.
Frozen Chronos-2 supplies the marginal predictive quantiles from
the observed history; these are converted to quantile functions
using Appendix~\ref{app:quantiles}.
The channel and temporal relations are estimated from the history
only, yielding $R_{\mathrm{ch}}[1,2]=0.8875$ and $\rho=0.9571$.
All three coupling methods share these marginal quantile functions
and a common base-normal tensor to generate $N=100$ sample paths.

The figure displays channels $1$ and $2$ as A and B, respectively,
and sample-path indices $0$ and $1$, using zero-based indexing.
It shows the last 96 history steps and all 48 forecast horizons.
Values are standardized using each channel's full pre-forecast
history.
The realization and displayed channel and path indices are fixed
without selection using future observations or method performance.
The figure illustrates the coupling mechanism rather than
benchmark forecast accuracy.

\section{Deterministic Median Point-Forecast Reference}
\label{app:median-reference}

The primary experiments compare alternative ways of coupling frozen
coordinate-wise predictive marginals into probabilistic multivariate
scenarios. As an additional reference, we ask a different question:
how do these probabilistic scenario forecasts compare with using only
the coordinate-wise predictive median as a single deterministic future?

This comparison is not a coupling ablation. The Median forecast is a
point-mass prediction, whereas IID, Temporal, Channel, Channel--Time,
and Schaake produce probabilistic scenario sets. We therefore report
Median only as a deterministic point-forecast reference and never use it
as the denominator of the IID-relative coupling skill reported in the
main experiments.

\subsection{Construction}
\label{app:median-construction}

For every channel $d$ and forecast horizon $h$, we define the deterministic
median trajectory as
\[
M_{d,h}
=
Q_{d,h}(0.5),
\]
where $Q_{d,h}$ is the same inverse quantile function used by the
probabilistic scenario methods.

The cached backbone outputs contain the quantile level $0.5$ explicitly.
As in the main experiments, predicted quantile knots are first monotonically
rearranged within each coordinate before constructing $Q_{d,h}$. Hence the
primary Median reference corresponds to the median of the same rearranged
marginal distribution from which all probabilistic scenarios are constructed.

The deterministic trajectory is evaluated using exactly the same
history-based standardization, realized future, and Variogram pair subsets
as the probabilistic methods. For implementation consistency, $M$ is
replicated $N=10$ times and passed through the existing scoring functions.

Because all replicated trajectories are identical, the pairwise spread term
in Energy Score vanishes, giving
\[
\mathrm{ES}_{\mathrm{Median}}
=
\left\|
\operatorname{vec}(M)
-
\operatorname{vec}(Y)
\right\|_2.
\]

Likewise, for the global-mean functional
\[
G(Y)
=
\frac{1}{DH}
\sum_{d=1}^{D}
\sum_{h=1}^{H}
Y_{d,h},
\]
the predictive distribution is a point mass at $G(M)$, and therefore
\[
\mathrm{CRPS}_{\mathrm{Mean,Median}}
=
\left|
G(M)-G(Y)
\right|.
\]

For the global maximum
\[
X(Y)
=
\max_{d,h}Y_{d,h},
\]
we similarly obtain
\[
\mathrm{CRPS}_{\mathrm{Max,Median}}
=
\left|
X(M)-X(Y)
\right|.
\]

Thus, the Median reference is scored using exactly the same metrics as the
probabilistic forecasts, but it contains no predictive spread.

\subsection{Comparison on the primary common support}
\label{app:median-primary}

Table~\ref{tab:median-primary} compares the Median reference with the five
probabilistic scenario constructions on the same evaluation set as the
primary controlled benchmark.

\begin{table*}[t]
\centering
\small
\setlength{\tabcolsep}{5pt}
\caption{
Dataset-equal absolute scores of the deterministic coordinate-wise Median
trajectory and the probabilistic scenario forecasts. Median is a deterministic point-forecast reference. All metrics are lower-is-better.
}
\label{tab:median-primary}
\begin{tabular}{llrrrrr}
\toprule
Type
& Method
& Energy
& Ch.-VS
& Time-VS
& Mean CRPS
& Max CRPS \\
\midrule

Point forecast
& Median
& 12.185
& 137.677
& 305.320
& 0.259
& 1.816 \\

\midrule

\multirow{5}{*}{Probabilistic}
& IID
& 8.910
& 121.082
& 217.623
& 0.232
& 1.551 \\

& Temporal
& 8.830
& 121.062
& 198.899
& 0.206
& 1.346 \\

& Channel
& 8.906
& 112.819
& 217.602
& 0.225
& 1.537 \\

& Channel--Time
& 8.805
& 112.781
& 199.096
& 0.201
& 1.341 \\

& Schaake
& 8.801
& 111.717
& 202.239
& 0.210
& 1.387 \\

\bottomrule
\end{tabular}
\end{table*}

The probabilistic scenario forecasts obtain lower aggregate scores than the
deterministic Median reference. However, this comparison has a different
interpretation from the fixed-marginal coupling experiment. Moving from
Median to IID changes the forecast representation itself: IID retains
multiple levels of each marginal predictive distribution and therefore
represents predictive uncertainty, whereas Median collapses each marginal
to a single value.

In contrast, the comparisons from IID to Temporal, Channel, and
Channel--Time hold the finite empirical marginal sample set exactly fixed
and modify only the dependence among coordinates. We therefore use Median
to contextualize the value of probabilistic scenario forecasting, while
IID remains the appropriate baseline for isolating the contribution of
dependence reconstruction.

\subsection{Paired comparisons against Median}
\label{app:median-paired}

For completeness, we perform paired comparisons between each probabilistic
forecast and Median. For method $M$, define
\[
\Delta(M)
=
S(M)-S(\mathrm{Median}).
\]
Since every metric is lower-is-better, $\Delta<0$ favors the probabilistic
forecast. Table~\ref{tab:median-paired} reports the paired differences and
bootstrap intervals.

\begin{table*}[!htbp]
\centering
\scriptsize
\setlength{\tabcolsep}{4pt}
\caption{
Paired raw-score comparisons against the deterministic Median reference on
the 654-cell primary support. Negative $\Delta$ indicates a lower score than
Median. CI denotes the hierarchy-aware 95\% percentile bootstrap interval.
The displayed differences use cell means, while the intervals use the
nested weighting described in Appendix~\ref{app:bootstrap}.
}
\label{tab:median-paired}
\begin{tabular}{llrrl}
\toprule
Method
& Metric
& $\Delta$
& 95\% CI
& W/T/L \\
\midrule

\multirow{5}{*}{IID}
& Energy
& -3.275
& [-5.210, -1.684]
& 14/0/0 \\
& Ch.-VS
& -16.595
& [-33.469, -4.259]
& 11/0/3 \\
& Time-VS
& -87.697
& [-179.110, -19.946]
& 8/0/6 \\
& Mean CRPS
& -0.027
& [-0.041, -0.015]
& 14/0/0 \\
& Max CRPS
& -0.264
& [-0.557, 0.008]
& 10/0/4 \\

\addlinespace

\multirow{5}{*}{Temporal}
& Energy
& -3.355
& [-5.272, -1.771]
& 14/0/0 \\
& Ch.-VS
& -16.615
& [-33.524, -4.276]
& 11/0/3 \\
& Time-VS
& -106.421
& [-199.010, -37.710]
& 14/0/0 \\
& Mean CRPS
& -0.053
& [-0.081, -0.032]
& 14/0/0 \\
& Max CRPS
& -0.469
& [-0.712, -0.264]
& 14/0/0 \\

\addlinespace

\multirow{5}{*}{Channel}
& Energy
& -3.279
& [-5.214, -1.686]
& 14/0/0 \\
& Ch.-VS
& -24.858
& [-47.066, -7.607]
& 11/0/3 \\
& Time-VS
& -87.718
& [-178.980, -20.081]
& 8/0/6 \\
& Mean CRPS
& -0.034
& [-0.051, -0.020]
& 14/0/0 \\
& Max CRPS
& -0.279
& [-0.570, -0.008]
& 10/0/4 \\

\addlinespace

\multirow{5}{*}{Channel--Time}
& Energy
& -3.380
& [-5.313, -1.782]
& 14/0/0 \\
& Ch.-VS
& -24.896
& [-47.134, -7.610]
& 11/0/3 \\
& Time-VS
& -106.225
& [-198.580, -37.589]
& 14/0/0 \\
& Mean CRPS
& -0.058
& [-0.090, -0.033]
& 14/0/0 \\
& Max CRPS
& -0.475
& [-0.712, -0.271]
& 14/0/0 \\

\addlinespace

\multirow{5}{*}{Schaake}
& Energy
& -3.384
& [-5.357, -1.760]
& 14/0/0 \\
& Ch.-VS
& -25.960
& [-48.820, -8.039]
& 11/0/3 \\
& Time-VS
& -103.081
& [-189.300, -37.979]
& 13/0/1 \\
& Mean CRPS
& -0.049
& [-0.070, -0.031]
& 14/0/0 \\
& Max CRPS
& -0.428
& [-0.690, -0.208]
& 14/0/0 \\

\bottomrule
\end{tabular}
\end{table*}

\subsection{Results}
\label{app:median-gaussian-support}

Table~\ref{tab:median-gaussian-support} compares Median with the Gaussian
coupling methods on the full evaluation set, without requiring Schaake
availability.

\begin{table}[!htbp]
\centering
\small
\setlength{\tabcolsep}{4.5pt}
\caption{
Dataset-equal absolute scores of Median and the Gaussian scenario methods. All metrics are lower-is-better.
}
\label{tab:median-gaussian-support}
\begin{tabular}{lrrrrr}
\toprule
Method
& Energy
& Ch.-VS
& Time-VS
& Mean CRPS
& Max CRPS \\
\midrule

Median
& 12.398
& 138.364
& 313.699
& 0.280
& 1.893 \\

IID
& 9.062
& 121.180
& 220.727
& 0.252
& 1.596 \\

Temporal
& 8.983
& 121.157
& 202.200
& 0.224
& 1.394 \\

Channel
& 9.059
& 113.102
& 220.699
& 0.243
& 1.582 \\

Channel--Time
& 8.957
& 113.055
& 202.388
& 0.217
& 1.388 \\

\bottomrule
\end{tabular}
\end{table}

\clearpage
\section{Fixed Temporal Correlation Controls}
\label{app:temporal-controls}

We compare the historical temporal coefficient with the fixed values
$\rho\in\{0,0.5,0.9,0.99\}$ at $N=100$, using native multivariate
inference for Chronos-2, TimesFM-3, and TiRex-2. Frozen marginal forecasts
are reused without additional backbone inference. Within each family,
only $R_{\mathrm{time}}[h,h']=\rho^{|h-h'|}$ changes: Temporal uses
$R_{\mathrm{ch}}=I_D$, while Channel--Time uses the same historical
$R_{\mathrm{ch}}$ across all five conditions. Every condition shares
the coordinate-wise empirical marginal sample sets and base Gaussian
draws. Energy and Time-VS are averaged over 20 sampling seeds per cell.
Table~\ref{tab:temporal-controls} includes all cases available to the Gaussian
methods, including those for which Schaake is unavailable.

\begin{table}[htbp]
\centering
\small
\setlength{\tabcolsep}{6pt}
\caption{IID-relative skill (\%) for temporal-correlation controls at
$N=100$, with all three backbones using
native multivariate inference. Ratios are formed from mean cell scores
within each dataset--backbone pair, then averaged equally over backbones
and datasets. Positive values indicate lower scores than IID.}
\label{tab:temporal-controls}
\begin{tabular}{lrrrr}
\toprule
& \multicolumn{2}{c}{Temporal} & \multicolumn{2}{c}{Channel--Time} \\
\cmidrule(lr){2-3}\cmidrule(lr){4-5}
$\rho$ & Energy & Time-VS & Energy & Time-VS \\
\midrule
History & 2.64 & 21.52 & 2.51 & 21.49 \\
0       & 0.00 & 0.00  & 0.02 & -0.02 \\
0.5     & 0.92 & 7.26  & 1.02 & 7.24 \\
0.9     & 2.42 & 20.27 & 2.52 & 20.25 \\
0.99    & 2.59 & 14.64 & 2.23 & 14.65 \\
\bottomrule
\end{tabular}
\end{table}

The fixed coefficient $0.9$ recovers most of the observed Time-VS
improvement. Historical estimation adds approximately $1.26$ and
$1.24$ percentage points for Temporal and Channel--Time, respectively,
but both paired intervals include zero (Table~\ref{tab:temporal-controls-paired}).

\begin{table}[htbp]
\centering
\small
\setlength{\tabcolsep}{4pt}
\caption{Paired Time-VS skill differences, History minus fixed $\rho_c$,
in percentage points. Positive values favor History. The first estimate
uses the headline aggregation; the nested estimate and percentile 95\%
interval use the hierarchical bootstrap described below.}
\label{tab:temporal-controls-paired}
\begin{tabular}{llrrl}
\toprule
Family & $\rho_c$ & Skill difference & Nested estimate & 95\% interval \\
\midrule
Temporal & 0.9  & 1.2556 & 1.2743 & $[-1.497,3.694]$ \\
Channel--Time & 0.9  & 1.2432 & 1.2613 & $[-1.534,3.636]$ \\
Temporal & 0.99 & 6.8859 & 6.8687 & $[0.4865,16.45]$ \\
Channel--Time & 0.99 & 6.8440 & 6.8266 & $[0.4435,16.28]$ \\
\bottomrule
\end{tabular}
\end{table}

The intervals resample datasets, items within datasets, and origins
within items using 10,000 paired draws after averaging sampling seeds.
Their nested point estimates average origins within items and then
items, whereas headline scores use plain cell means within each
dataset--backbone pair. These weighting conventions are distinguished
in Table~\ref{tab:temporal-controls-paired}; intervals are not adjusted
for multiple comparisons.

\section{Paired Comparisons between Coupling Baselines}
\label{app:paired-couplings}

We compare Channel--Time directly with Temporal and Schaake using the
stored fixed-marginal scores on the same common support as the main
benchmark. All three backbones use native multivariate inference; no
new forecasts or sampled paths are introduced. Table~\ref{tab:paired-couplings-primary}
reports the primary Energy and global-max CRPS contrasts at $N=100$.

\paragraph{Contrast and uncertainty.}
The skill difference is
\[
\Delta_{\mathrm{skill}}(A,B)
=\frac{1}{|\mathcal D|}\sum_{d\in\mathcal D}
  \frac{1}{3}\sum_b\bigl[\operatorname{Skill}_{d,b}(A)
                         -\operatorname{Skill}_{d,b}(B)\bigr],
\]
where $A$ is Channel--Time and $B$ is Temporal or Schaake.
Each skill uses the IID score of the same dataset--backbone pair as its
denominator. Positive differences, measured in percentage points (pp),
favor Channel--Time.

Intervals follow the matched hierarchical bootstrap described in
Appendix~\ref{app:bootstrap}. They are individual percentile intervals
without multiplicity correction. An interval containing zero does not
establish equivalence.

\begin{table}[htbp]
\centering
\small
\setlength{\tabcolsep}{5pt}
\caption{Primary paired comparisons at $N=100$ under exactly fixed
empirical marginals. Entries are Channel--Time minus the comparator's
IID-relative skill, in percentage points; positive values favor
Channel--Time. Intervals are individual 95\% paired bootstrap intervals.}
\label{tab:paired-couplings-primary}
\begin{tabular}{llrl}
\toprule
Comparator & Metric & $\Delta_{\mathrm{skill}}$ (pp) & 95\% CI \\
\midrule
Temporal & Energy & $-0.133$ & $[-0.637,\,0.320]$ \\
Temporal & Global-max CRPS & $1.451$ & $[-3.173,\,3.945]$ \\
\addlinespace
Schaake & Energy & $0.722$ & $[0.295,\,1.294]$ \\
Schaake & Global-max CRPS & $4.734$ & $[0.074,\,12.918]$ \\
\bottomrule
\end{tabular}
\end{table}

\paragraph{Comparison scope.}
The Energy and global-max CRPS intervals do not establish an additional
advantage of Channel--Time over Temporal. Against Schaake, both primary
skill intervals favor Channel--Time. These findings concern the reported
metrics and normalized contrasts and do not establish uniform dominance
over historical rank coupling.

\section{Dataset-Level Effects}
\label{app:dataset-effects}

We summarize the same fixed-marginal scores by dataset, averaging the
three backbone-specific IID-relative skills with equal weight.
Table~\ref{tab:dataset-ct-n100} reports all five metrics for Channel--Time
at $N=100$; Table~\ref{tab:dataset-wtl-n100} gives win/tie/loss counts for
each method and the two direct comparisons. Counts use the signs of
unrounded point values: positive, exactly zero, and negative skill
(or skill difference), respectively. They are descriptive counts, not
per-dataset significance decisions. 
\begin{table}[htbp]
\centering
\scriptsize
\setlength{\tabcolsep}{4pt}
\caption{Dataset-level IID-relative skill (\%) of Channel--Time at $N=100$
under exactly fixed empirical marginals. Values average skill equally
across the three native-multivariate backbones. Negative entries indicate
higher scores than IID; they are not omitted from the overall mean.
Mean and Max denote CRPS of the standardized global-mean and global-max
functionals.}
\label{tab:dataset-ct-n100}
\begin{tabular}{lrrrrr}
\toprule
Dataset & Energy & Ch.-VS & Time-VS & Mean & Max \\
\midrule
ETT 15T & 0.679 & 14.476 & 8.009 & 21.730 & 29.119 \\
ETT 1D & 0.118 & -0.593 & 4.452 & 9.150 & 17.591 \\
ETT 1H & 0.280 & 10.906 & 4.899 & 14.467 & 25.691 \\
Bizitobs L2C 1H & -0.264 & 6.708 & 29.746 & 15.932 & 54.474 \\
Boomlet 963 & 1.224 & 7.221 & 5.418 & 22.453 & 12.000 \\
FRED-MD & 4.918 & 0.503 & 34.223 & 13.502 & 29.192 \\
FRED-QD & 9.265 & -3.929 & 62.304 & 11.435 & 17.583 \\
GVAR & 0.634 & 1.845 & 30.657 & 6.876 & 38.682 \\
Jena Weather 1H & 1.068 & 13.673 & 27.195 & 7.512 & 13.364 \\
UCI Air Quality 1D & 0.547 & 16.387 & 4.615 & -98.261 & 28.182 \\
UK COVID 1D, cumulative & 7.415 & 3.068 & 34.086 & 12.230 & 46.738 \\
UK COVID 1D, new & 1.579 & 1.614 & 11.318 & 13.734 & 21.167 \\
UK COVID 1W, cumulative & 2.991 & 2.857 & 12.379 & 18.276 & 11.635 \\
UK COVID 1W, new & 4.778 & -9.118 & 33.221 & 22.842 & 19.593 \\
\midrule
Dataset-equal mean & 2.517 & 4.687 & 21.609 & 6.563 & 26.072 \\
\bottomrule
\end{tabular}
\end{table}

\begin{table}[htbp]
\centering
\scriptsize
\setlength{\tabcolsep}{4pt}
\caption{Dataset win/tie/loss counts at $N=100$ from the signs of
backbone-averaged point estimates across 14 datasets. The first four
columns compare each method with IID; the last two compare Channel--Time
(CT) with Temporal (T) and Schaake (S), using skill differences.
These counts are not significance tests.}
\label{tab:dataset-wtl-n100}
\begin{tabular}{lcccccc}
\toprule
& \multicolumn{4}{c}{Method versus IID}
& \multicolumn{2}{c}{Direct comparison} \\
\cmidrule(lr){2-5}\cmidrule(lr){6-7}
Metric & Temporal & Channel & CT & Schaake & CT vs T & CT vs S \\
\midrule
Energy & 14/0/0 & 9/0/5 & 13/0/1 & 10/0/4 & 9/0/5 & 13/0/1 \\
Ch.-VS & 7/0/7 & 11/0/3 & 11/0/3 & 11/0/3 & 11/0/3 & 7/0/7 \\
Time-VS & 14/0/0 & 6/0/8 & 14/0/0 & 13/0/1 & 5/0/9 & 12/0/2 \\
Global-mean CRPS & 13/0/1 & 12/0/2 & 13/0/1 & 13/0/1 & 8/0/6 & 11/0/3 \\
Global-max CRPS & 14/0/0 & 13/0/1 & 14/0/0 & 13/0/1 & 7/0/7 & 10/0/4 \\
\bottomrule
\end{tabular}
\end{table}

\clearpage
\end{document}